\documentclass[letterpaper]{article} 
\usepackage{aaai2027}  

\usepackage[hyphens]{url}  
\usepackage{graphicx} 
\usepackage{amsfonts} 
\usepackage{bm} 
\usepackage{dsfont}
\usepackage{amsmath} 
\usepackage{natbib}  
\usepackage{caption} 
\usepackage{booktabs}
\usepackage{multirow}

\DeclareMathOperator{\NLL}{NLL}
\DeclareMathOperator{\softmax}{softmax}

\title{ProtoGestalt: Interpretable Prototypes for Facial Phenotyping and Diagnosis of Rare Genetic Disorders}
\author {
    Ranga Baminiwatte\textsuperscript{\rm 1,\rm 2},
    Alexander Hustinx\textsuperscript{\rm 3},
    Fabio Hellmann\textsuperscript{\rm 4},
    Aaron J. Masino\textsuperscript{\rm 1,\rm 2}\corresponding
}
\affiliations {
    \textsuperscript{\rm 1}School of Computing, Clemson University, USA\\
    \textsuperscript{\rm 2}Institute for Human Genetics, Clemson University, USA\\
    \textsuperscript{\rm 3}Institute for Genomic Statistics and Bioinformatics, Rheinische Friedrich-Wilhelms Universität Bonn, Germany\\
    \textsuperscript{\rm 4}Human-centered Artificial Intelligence Lab, Augsburg University, Germany\\

}

\begin{document}

\maketitle

\begin{abstract}

Rare genetic disorders are difficult to diagnose due to their diversity and overlapping clinical features. Facial morphology can provide diagnostic clues, yet most diagnostic models offer limited interpretability and are not designed to support broad screening for undiagnosed disease. We present ProtoGestalt, an interpretable prototype-based framework for facial phenotyping, rare-disease screening, and syndrome identification. The method extends prototype learning by constraining learned visual features to select facial regions, allowing predictions to be explained through localized and additive evidence. We assess ProtoGestalt's ability to distinguish syndromic from unaffected faces on multiple control datasets and class prevalence values. We further evaluate it's diagnostic performance on syndromes observed during training. The learned prototype-based representations provide significant separation between syndromic and case-matched control images and retain useful performance as the proportion of controls increases. The framework also demonstrates the ability to differentiate among syndromes. Although reasonably accurate specialized models for facial classification of rare disorders already exist, ProtoGestalt offers an interpretable alternative that links predictions to anatomically meaningful facial regions, potentially supporting clinical decision-making and the discovery of phenotype–gene associations.

\end{abstract}


\section{Introduction}

Rare genetic disorders affect more than 6\% of the global population. There are an estimated 7,000–10,000 known rare diseases, approximately 80\% of which have a genetic etiology \cite{Marwaha2022}, while fewer than 10\% have an approved therapy \cite{Halley2022}. Due to the large number of disorders, extensive phenotypic heterogeneity, and overlapping clinical presentations, the median time to diagnosis is 4-7 years \cite{Heuyer2017, Molster2016}. 

A necessary step toward a rare disease diagnosis is referral for genetic testing by a front-line clinician. This referral, however, is usually substantially delayed relative to symptom onset because clinicians do not consider rare diseases in their initial assessments. This is partly because two-thirds of clinicians significantly underestimate the prevalence of rare diseases, considering them to be 50 to 500 times rarer than the standard definitions used in the U.S. and Europe \cite{Rohani-Montez2023}. Additionally, rare diseases exhibit complex, variable phenotypic presentations that may overlap with those of more common conditions, further delaying evaluation. Indeed, prior studies indicate that many clinicians may not recognize rare disease phenotypes and report a lack of knowledge of rare disease symptoms as a barrier to diagnosis \cite{Rohani-Montez2023, Budych2012}. Even after rare conditions are considered, the existence of thousands of phenotypically complex rare conditions, many with overlapping features, presents a significant challenge to determining a specific diagnosis.

Computational methods could assist with early recognition of individuals with undiagnosed rare disease (i.e., those for whom rare disease has yet to be considered in diagnostic planning) and subsequent disease specific diagnosis. Surprisingly, there has been only limited work on identifying patients with undiagnosed rare diseases, with most existing studies focusing on screening for a single disease. For example, Bhasuran et al. developed a machine learning pipeline to identify patients with acute hepatic porphyria \cite{Bhasuran2025}. While potentially useful, this approach would require a separate model for each rare disease, making this paradigm impractical as a general screening strategy. In contrast, the \emph{ThinkRare} model uses a rule-based approach applied to structured EHR data to identify pediatric patients with any undiagnosed multisystem rare genetic disease who are eligible for exome sequencing \cite{Ediae2025}. However, the rule-based design encodes fixed expert-defined clinical eligibility criteria, limiting adaptability across institutions with different documentation conventions or patient populations.

Many algorithmic models, including \emph{Xrare} \cite{Li2019}, \emph{LIRICAL} \cite{Robinson2020}, \emph{PhenoBrain} \cite{Mao2025}, and \emph{SHEPHERD} \cite{Alsentzer2025}, have been developed to infer a patient's specific rare disease diagnosis. These methods rely on detailed phenotypic information about the patient, typically using a controlled vocabulary such as the Human Phenotype Ontology \cite{Khler2021}, and, optionally, genetic sequence data. This phenotype labeling requirement poses a substantial challenge as ontologies such as the HPO are complex. Indeed, the current HPO version (v2026-06-23) contains over 20,000 terms. Additionally, it includes many subjective terms that are difficult for non-experts to apply in practice. For example, the term midline nasal groove is described as “an abnormal groove on the midline of the nose that may extend to the nasal tip.” Finally, none of these systems have been adapted to the problem of identifying individuals with an undiagnosed rare disease.

To address these limitations, \emph{DeepGestalt} \cite{Gurovich2019} and \emph{GestaltMatcher} \cite{Hsieh2022} use deep learning methods on facial images for rare disease diagnosis. The facial gestalt, or overall facial morphology, is often a characteristic clinical feature indicating presence of an undiagnosed rare disease, with certain features associated with specific diagnoses \cite{Kirchhoff2025}. Approximately 30–40\% of genetic disorders include facial dysmorphology features \cite{Ferry2014}. Both \emph{DeepGestalt} and \emph{GestaltMatcher} are end-to-end models and neither provides a mechanism for ascribing specific, clinically interpretable phenotypic features to individual patients or diseases. As with the aforementioned rare disease diagnostic models, these models have not been adapted to support undiagnosed rare disease screening. 

In this paper, we introduce \emph{ProtoGestalt}, a prototype learning model to support both identification of individuals with undiagnosed rare diseases and disease diagnosis for individuals suspected of having a rare disease. \emph{ProtoGestalt} is an extension of \emph{PIP-Net} \cite{nauta2023} which learns latent-space prototypes that represent characteristic portions of objects, in our case facial regions. We extend this approach by enforcing anatomical constraints on the prototype learning process using facial landmarks. The primary advantage of our approach is that the model predictions are inherently explainable through visualization of the matching prototypes for a given input. Additionally, prototypes that are consistently present across individuals with the same condition can potentially be used to objectively characterize the disease, which may help support rare disease research. We make the following contributions: (1) develop \emph{ProtoGestalt}, an anatomy constrained extension of the \emph{PIP-Net} model that learns prototypic facial dysmorphology characteristics associated with rare disease and performs disease diagnosis for common rare disorders; (2) present the novel task of determining if an individual has an undiagnosed rare disease or is unaffected by a rare disease based on prototypical representations for a given input facial image ; (3) extend \emph{ProtoGestalt} to infer if an input image is of an individual affected or not affected by a rare disease. 

\section{Related Work}
\subsection{Rare Disease Diagnosis via Facial Images} Automated facial analysis systems have demonstrated strong performance for rare disease diagnosis. \emph{DeepGestalt} is a 
 convolutional neural network trained on over 17,000 patient images spanning 216 syndromes. It achieved 91\% top-10 accuracy in syndrome identification. This approach was extended with \emph{GestaltMatcher}, which encodes facial images into a 320-dimensional Clinical Face Phenotype Space and enables patient-to-patient matching, supporting recognition of ultra-rare syndromes not represented in the training set. It was later improved with updated CNN backbone models, model ensembles, and test time augmentation, ultimately achieving top-1 and top-5 disease diagnostic performances of 49\% and 72\%, respectively, across 204 disorders \cite{Hustinx2022}. The internal representations learned by these models are not interpretable, limiting their utility for objective phenotype characterization independent of a diagnostic label.

\subsection{Synthetic Rare Disease Facial Image Generation} To address data scarcity and privacy constraints that impede training AI systems, some studies have explored generative AI to produce synthetic facial images with disorder-specific dysmorphology. Kirchhoff et al. used a conditional StyleGAN architecture to generate portraits for 20 disorders \cite{Kirchhoff2025}. Benichou et al. applied diffusion models with syndrome-specific model fine-tuning to produce ethnically balanced synthetic faces for 10 syndromes, demonstrating improved classifier performance when synthetic images supplemented real training data \cite{Benichou2025}. However, both approaches generate whole-face images intended to augment training data for diagnostic models and do not produce structured, reusable phenotypic descriptors.

\subsection{Prototypical Networks}
Prototype-based interpretable networks are designed to explain predictions by comparing image regions with learned representative patches. For instance, ProtoPNet \cite{Chen2019} classifies images by comparing parts of the input to learned "prototypes" (e.g., specific features like a bird's beak), to provide understandable class-based explanations. PIP-Net \cite{nauta2023} improved prototype organization and interpretability by learning sparse, human-aligned prototypes through self-supervision and a non-negative classifier. These models have also been explored in explaining medical imaging tasks \cite{yazdani2025}, \cite{wei2024}, \cite{nauta2023_b}. However, prototype-based networks have not been applied to rare-disease facial phenotyping, and existing methods do not enforce anatomically consistent prototype locations. 

\section{Data \& Methodology}
The Clemson University Institutional Review Board (IRB) reviewed and approved the study under expedited review (IRB2025-1191), and a waiver of informed consent was granted.

\begin{figure*}[t]
    \centering  \includegraphics[width=1.0\textwidth, keepaspectratio]{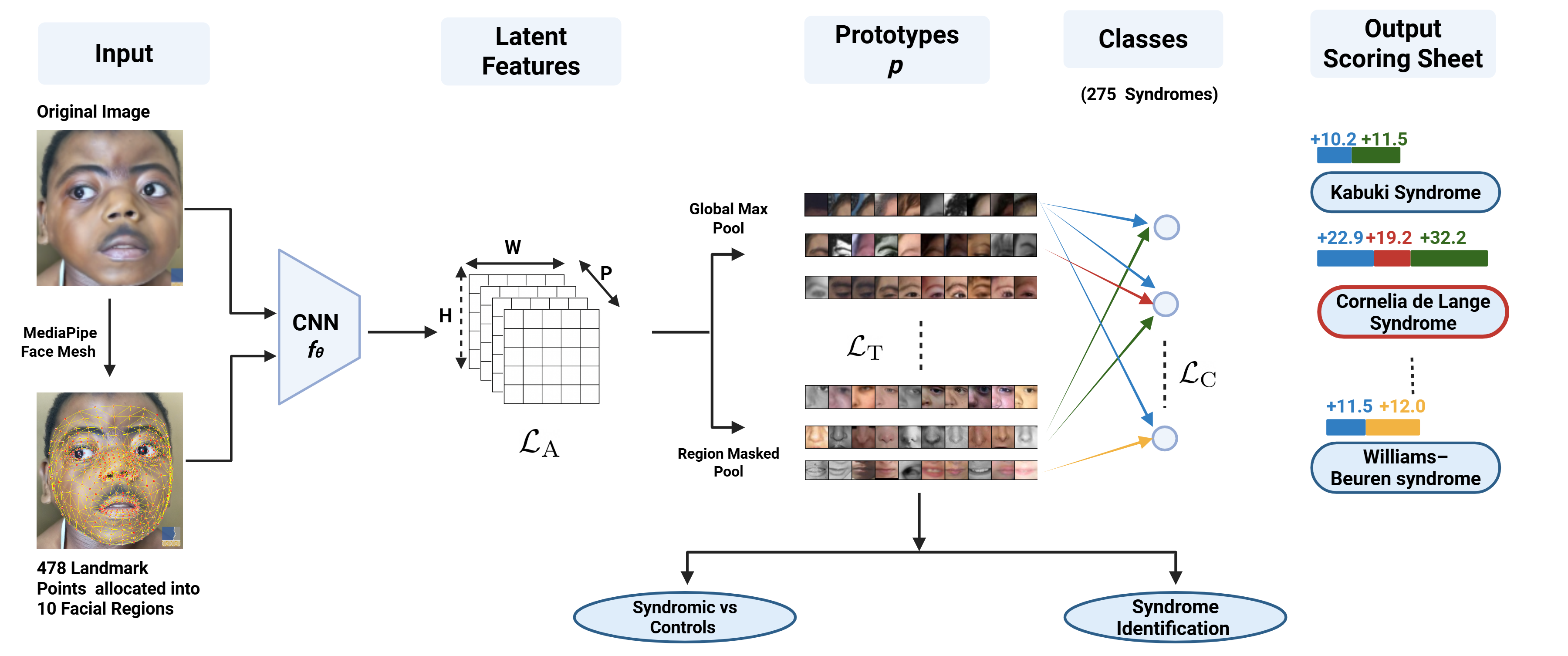}
    \caption{\textbf{Overview   of the ProtoGestalt framework.}
    Given an input facial image $\mathbf{x}$, a CNN backbone $f_{\theta}$ produces $P$ spatial prototype activation maps of size $H \times W$. Pooling generates prototype similarity scores which are passed to a prediction head, $h_{\phi}$ for output inference. MediaPipe Face Mesh detects 478 facial landmarks, which define ten anatomical regions. Each prototype is assigned to one region, and its presence score is obtained using region-masked max pooling; global max pooling is used when the required landmarks are unavailable. The synthetic image shown in the illustration was generated with GestaltGAN \cite{Kirchhoff2025} for use in this publication.}
\label{fig:method}
\end{figure*}

\subsection{Data}
\subsubsection{Facial images of rare disorders}

We used the GestaltMatcher Database (GMDB) \cite{Hsieh2022} version 1.1.0 for model training and evaluation. GMDB contains 10,980 frontal images of 8,346 unique patients with 581 different syndromes. We followed the data splitting strategy proposed in \cite{Hustinx2022} and split the data across syndromes into two sets: GMDB-Frequent (syndromes with more than six patients) and GMDB-Rare (syndromes with six or fewer patients). There are 9,671 images of 7,294 patients with 275 syndromes in GMDB-Frequent. GMDB-Rare contains 1,309 images of 1,052 patients with 306 syndromes.	

We further split GMDB-Frequent into training, validation and test sets and used it to train, fine-tune and evaluate \emph{ProtoGestalt}. We used GMDB-Rare only for evaluation as it provides a syndrome-disjoint test set for assessing whether the learned prototypes generalize to facial images from syndromes that were not observed during training. This evaluation setting reflects clinical diagnosis of ultra rare conditions, where the underlying disorder may not be represented in training data. An overview of the dataset and experimental partitions is provided in Table~\ref{tab:GMDB_data}.

\begin{table}[t]
\centering
\small
\setlength{\tabcolsep}{4pt}
\begin{tabular}{@{}llrrr@{}}
\toprule
\textbf{Dataset} &
\textbf{Images} &
\textbf{Patients} &
\textbf{Syndromes} \\
\midrule
GMDB-Frequent
    & 9671
    & 7294
    & 275 \\

\quad Training
    & 7787            
    & 5855
    &  275 \\

\quad Validation
    &  1002       
    &  778
    &  275 \\

\quad Test
    & 882       
    & 661
    &  275 \\
\midrule
GMDB-Rare
    & 1309
    & 1052
    & 306 \\
\midrule[1.5pt]

GMDB-Total
    & 10980		
    & 8346
    & 581 \\
\bottomrule
\end{tabular}
\caption{Overview of GMDB (v. 1.1.0) dataset. Training and validation sets of GMDB-Frequent are used for training and fine-tuning prototypes. Test set of GMDB-Frequent and GMDB-Rare sets are used for evaluation.}
\label{tab:GMDB_data}
\end{table}

\subsubsection{Control datasets}

To evaluate the utility of the prototypes for distinguishing individuals with rare disorders from unaffected individuals (controls), we used two open facial-image datasets: CASIA-WebFace \cite{yi2014} and UTKFace \cite{zhang2017}. These datasets contain 494,414 and 23,708 images, respectively. UTKFace provides annotations for age, sex, and ethnicity. The GMDB dataset predominantly contains images of children (92.4\% $<18$ years of age). In contrast, only 17.9\% of the images in UTKFace depict individuals under 18. To account for this imbalance and further evaluate our method, we constructed a matched control cohort from UTKFace following the approach described by Pantel et al.~\cite{pantel2020}. Each test image from GMDB was matched with an image of an individual without a rare disorder based on age, sex, and ethnicity.
We cropped and aligned all images based on image-preprocessing method proposed in \cite{Hsieh2024}.


\subsection{Methods}
\subsubsection{PIP-Net}
\label{sec:pipnet}

We based our model on PIP-Net \citep{nauta2023}, a prototype-based classifier in which prototypes are learned through self-supervision. Given an input image $\mathbf{x}\in\mathbb{R}^{3\times S\times S}$, a backbone CNN $f_\theta$ followed by a softmax-based prototype head $h_\phi$, the model produces $P$ spatial prototype activation maps:

\begin{equation}
\mathbf{z}
=
h_\phi\!\big(f_\theta(\mathbf{x})\big)
\in\mathbb{R}^{P\times H\times W},
\label{eq:protomaps}
\end{equation}

\noindent The softmax, $\sum_{d=1}^{P} z_{d,i,j}=1$, is applied across the prototype dimension so that at each spatial location $z_{d,i,j}\in[0,1]$ represents the probability that the local image patch corresponds to prototype $d$.

PIP-Net converts each activation map into a scalar presence score using
\emph{global max pooling} over $z_{d,:,:}$ for each $d$ such that:
\begin{equation}
p_d
=
\max_{(i,j)} z_{d,i,j}.
\label{eq:globalpool}
\end{equation}
The score $p_d$ therefore represents the maximum probability that prototype $d$ matches some portion of the image. The resulting \emph{prototype presence vector}, $\mathbf{p}=(p_1,\ldots,p_P)$, is passed to a sparse, non-negative linear classifier. As the classifier weights are non-negative, each class score is an additive combination of the detected prototypes. The combination of class scores acts as a scoring sheet to interpret the prediction.

\subsubsection{Region-Masked Prototypes}
\label{sec:rmpp}
A limitation of the global max pooling used in \emph{PIP-Net} is that it ignores spatial location. A prototype can contribute to the prediction regardless of where its strongest activation occurs, and the maximizing location may change across images. This is problematic for facial phenotyping, where a prototype should represent a feature from a consistent anatomical region. For example, a prototype intended to capture a periorbital feature should only activate on that region.

To enforce anatomically consistent prototype evidence, we replace global max pooling with a max-pooling operator restricted to a predefined facial region. Each prototype is assigned to one of ten ($R=10$) semantic regions, such as the \emph{left eye}, \emph{nose}, or \emph{mouth}. The regions are listed in Table~\ref{tab:regions} and are defined using fixed sets of MediaPipe Face Mesh\cite{grishchenko2020} landmark indices $\mathcal{I}_g$ where $g$ indexes facial regions.

For a given image, let $\bm{\ell}^{m}\in[0,1]^2$ denote the normalized coordinate of landmark $m$, and let $v_m\in\{0,1\}$ indicate whether that landmark is valid ($\bm{\ell}^{m}$ is not missing and lies inside the image). We create a binary mask for region $g$ at the resolution of the prototype activation maps:

\begin{equation}
M^{g}_{i,j}
=
\mathds{1}\!\left[
\exists\,m\in\mathcal{I}_g:
v_m=1
\;\wedge\;
\left\lVert
\mathbf{c}_{i,j}-\bm{\ell}^{m}
\right\rVert_2
\le \rho
\right],
\label{eq:mask}
\end{equation}
where $\mathbf{c}_{i,j}\in[0,1]^2$ is the normalized center coordinate of grid cell $(i,j)$. Thus, a cell belongs to the mask if it lies within radius $\rho$ of at least one valid landmark assigned to that region.

The radius $\rho$, expressed as a fraction of the image side length, controls the spatial extent of each mask. Small values produce tightly localized masks around the landmarks, whereas sufficiently large values cause the masks to cover the full activation grid. In the latter case, the method reduces to standard global max pooling.

We distribute prototypes approximately evenly across the $R$ regions using a
round-robin assignment:
\begin{equation}
r(d)
=
1+\big((d-1)\bmod R\big),
\qquad d\in\{1,\ldots,P\},
\label{eq:assign}
\end{equation}
where $r(d)$ denotes the region assigned to prototype $d$. Regions are indexed
from $1$ to $R$ in the order shown in Table~\ref{tab:regions}.

The presence score of prototype $d$ is then computed only within its assigned
region:
\begin{equation}
p_d =
\begin{cases}
\displaystyle
\max_{(i,j):\,M^{r(d)}_{i,j}=1} z_{d,i,j},
&
\text{if }
\displaystyle\sum_{i,j}M^{r(d)}_{i,j}>0,
\\[1.4ex]
\displaystyle
\max_{(i,j)} z_{d,i,j},
&
\text{otherwise.}
\end{cases}
\label{eq:maskpool}
\end{equation}
The second case provides a fallback to global pooling when the assigned
region has an empty mask, for example because the required landmarks were not
detected.
Whenever a valid regional mask is available, activations outside that mask
cannot affect $p_d$, even if they are larger than every activation inside the
region. Consequently, prototype $d$ can contribute evidence only from its
assigned anatomical region. This is a hard structural constraint, in contrast
to soft regularization terms that merely penalize off-region activations
without preventing them from contributing to the prediction. The masking
operation introduces no learnable parameters and changes only the pooling
stage. A backbone pretrained using standard PIP-Net pooling can therefore be
fine-tuned directly with region-Masked pooling.

\begin{table}[t]
\centering
\begin{tabular}{@{}l l@{}}
\toprule
Region $g$ & Landmark indices $\mathcal{I}_g$ \\
\midrule
left eye    & 468, 469, 470, 471, 472, 33, 133, 159, 145 \\
right eye   & 473, 474, 475, 476, 477, 263, 362, 386, 374 \\
left brow   & 105, 66, 107, 55, 65 \\
right brow  & 334, 296, 336, 285, 295 \\
nose        & 1, 2, 4, 5, 6, 19, 94 \\
left cheek  & 50, 116, 117, 118, 205 \\
right cheek & 280, 345, 346, 347, 425 \\
mouth       & 13, 14, 61, 291, 0, 17, 78, 308 \\
chin        & 152, 175, 199, 148, 377 \\
forehead    & 10, 9, 151, 8, 107 \\
\bottomrule
\end{tabular}
\caption{The ten facial regions and their defining MediaPipe Face Mesh
landmark sets $\mathcal{I}_g$. MediaPipe provides 468 facial-mesh landmarks
and 10 additional iris landmarks, for a total of 478 points. Regions are
indexed in the order shown, and each prototype is assigned to one region
according to Eq.~\eqref{eq:assign}.}
\label{tab:regions}
\end{table}

\begin{figure}[t]
    \centering
    \includegraphics[width=1.0\columnwidth]{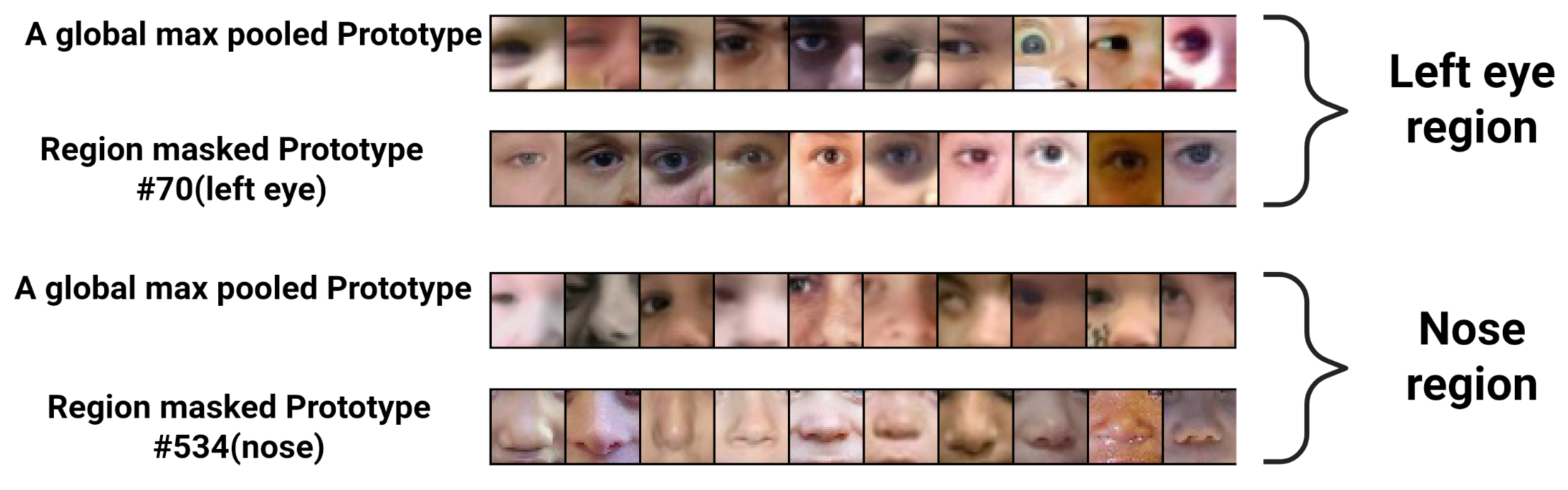}
    \caption{Visualizations of a global Max Pooled prototype and a region Masked prototype at a specific region. Fixing prototypes to specific regions mitigates the potential mirror effect of CNNs (Example: recognizing the right eye as a left eye). Furthermore, region Masked prototypes allow us to track which prototype corresponds to which facial region.}
    \label{fig1}
\end{figure}

\subsubsection{Training Objective}
\label{sec:loss}

We retain PIP-Net's self-supervised alignment and diversity objectives. The main training objective is
\begin{equation}
\mathcal{L}
=
\lambda_{\mathrm{A}}\mathcal{L}_{\mathrm{A}}
+
\lambda_{\mathrm{T}}\mathcal{L}_{\mathrm{T}}
+
\lambda_{\mathrm{C}}\mathcal{L}_{\mathrm{C}},
\end{equation}

The alignment loss $\mathcal{L}_{\mathrm{A}}$ encourages corresponding patches
from two augmented views of the same image to select the same prototype. The
diversity loss $\mathcal{L}_{\mathrm{T}}$ encourages all prototypes to be used
within a training batch. Their definitions are unchanged; region-Masked
pooling replaces global pooling wherever a pooled prototype score is required.

For classification, the pooled prototype scores are passed to a non-negative
linear layer:
\begin{equation}
\begin{aligned}
\mathbf{o}
&=
\mathbf{W}\mathbf{p},
\qquad
\mathbf{W}\in\mathbb{R}_{\ge 0}^{C\times P},
\\
\mathcal{L}_{\mathrm{C}}
&=
\NLL\!\left(
\log\softmax\!\left(
\log\!\left(1+\mathbf{o}^{\,\tau}\right)
\right),
y
\right),
\end{aligned}
\label{eq:cls}
\end{equation}
where $\mathbf{o}\in\mathbb{R}^{C}$ contains the class scores, $C$ is the
number of classes, $\tau \ge 1$ is a learnable scalar parameter, and $y$ is the ground-truth label. The exponentiation and logarithms are applied elementwise.

To obtain the prototypes, we trained PIPnet to classify 275 syndromes using GMDB-Frequent training data. We then used the prototypes (both self-supervised and region-Masked) to represent test images in the evaluations.

We obtain $M=478$ landmarks per image using MediaPipe Face Mesh and construct
the regional masks at the backbone feature-map resolution, $H=W=14$. At
inference time, region-masked pooling is used whenever valid landmarks are
available; if the relevant landmarks are unavailable, the fallback in
Eq.~\eqref{eq:maskpool} reduces the operation to standard PIP-Net global max
pooling. Every image had an input resolution of $S=112$. We set the mask radius to $\rho=0.1$, expressed as a fraction of the
normalized image, which defines the disk drawn around each landmark when
forming the binary region masks. All model hyperparameters were tuned using Optuna \cite{akiba2019} hyper parameter tuning framework.

\subsubsection{Evaluation}
\label{sec:eval}

For each image $\mathbf{x}$, we used the prototype-presence vector
$\mathbf{g}(\mathbf{x})\in\mathbb{R}^{P}$ as its representation, computed
using either global max pooling (Eq.~\eqref{eq:globalpool}) or region-Masked
pooling (Eq.~\eqref{eq:maskpool}).

We evaluated whether these representations could distinguish images of
individuals with genetic syndromes from control facial images. A binary
logistic-regression (LR) classifier modeled the probability that an image
belonged to the syndromic class as

\begin{equation}
P(y=1\mid\mathbf{x})
=
\sigma\left(
\boldsymbol{\beta}^{\top}\mathbf{g}(\mathbf{x})+b
\right),
\label{eq:lr}
\end{equation}

\noindent where $y=1$ denotes a GMDB image, $y=0$ denotes a control image, $\sigma$ is
the sigmoid function, $\boldsymbol{\beta}\in\mathbb{R}^{P}$ contains one
learned coefficient per prototype, and $b$ is the intercept.

For each pooling method, we trained two LR classifiers: one using
CASIA-WebFace controls and one using UTKFace controls, resulting in four
classifiers in total. Each classifier was trained on 2,000 randomly selected
GMDB-Frequent training images and 2,000 images from the corresponding control
dataset. The GMDB-Frequent images were sampled independently for the
CASIA-WebFace and UTKFace training configurations. Control images used for
evaluation were excluded from training.

The syndromic evaluation cohort combined the GMDB-Frequent test set
($n=882$) and GMDB-Rare ($n=1,309$), yielding 2,191 images. We constructed
three evaluation sets by pairing this cohort with approximately the same
number of held-out controls from CASIA-WebFace, UTKFace, and the matched
UTKFace cohort. The CASIA-WebFace-trained classifiers were evaluated using
CASIA-WebFace controls, whereas the UTKFace-trained classifiers were
evaluated separately using the general and matched UTKFace controls.

To more closely approximate a clinical screening setting, in which controls
are expected to substantially outnumber affected individuals, we also
evaluated robustness to class prevalence. We held the 2,191 syndromic images
fixed while progressively increasing the number of control images in the
test set, and reported Area Under the Precision-Recall Curve (AUPRC) as a function of the resulting class imbalance.

As a secondary task We evaluated the ability of the prototype representations to distinguish among
syndromes that were observed during training(GMDB-Frequent). We directly used the trained PIP-Net
classification head for this purpose. For an image $\mathbf{x}$ with class-score
vector $\mathbf{o}(\mathbf{x})\in\mathbb{R}^{C}$, the predicted class is

\[
\hat{y}(\mathbf{x})=\arg\max_{c} o_c(\mathbf{x}),
\]

For $N$ test images, top-$k$ accuracy is

\[
\mathrm{Acc}_k=\frac{1}{N}\sum_{i=1}^{N}
\mathbb{I}\!\left[y_i\in T_k(\mathbf{x}_i)\right].
\]
\noindent where $T_k(\mathbf{x})$ denotes the $k$ classes with the highest scores. We present results for $k \in [1,5,10]$

\section{Results}

\subsection{Syndromic-versus-control classification performance}
\begin{figure}[t]
    \centering
    \includegraphics[width=0.9\columnwidth]{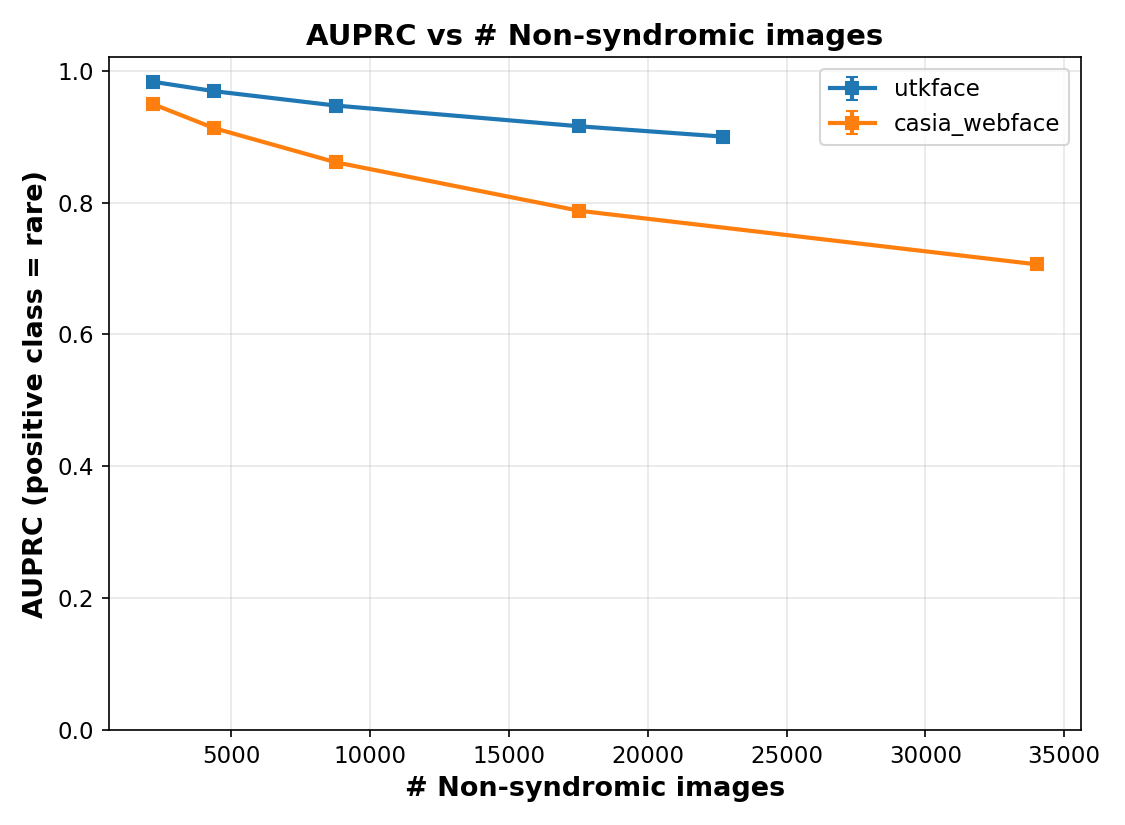}
    \caption{AUPRC values as the number of control images increases while the syndromic cohort remains fixed, evaluating LR model's (Global max pooled) robustness under increasing control-to-syndromic class imbalance and decreasing syndrome prevalence.AUPRC declined progressively yet remained above 0.70 at all evaluated imbalance levels.}
    \label{fig5}
\end{figure}

%
Under demographically matched controls, which isolate dysmorphology from age and dataset biases, prototype representations separated syndromic from unaffected faces at 0.85 accuracy(GMP) and 0.73(RM). Against unmatched UTKFace controls both rise sharply (0.94, 0.86), a gap we attribute to the age imbalance between cohorts. Region masking produced higher precision and specificity for CASIA-WebFace and general UTKFace controls, but lower recall and F1. As the number of controls increased, AUPRC declined, yet remained above 0.70 at all evaluated imbalance levels (Figure~\ref{fig5}). All model variants were deployed using ConvNeXt26 \cite{gao2024} as the backbone CNN($f_\theta$).


\begin{table}[t]
\centering
\begin{tabular}{c|cccccc}
\hline
 Pooling & Controls &Acc &PPV &Sen&Spec&F1 \\
\hline
 GMP & C-WF &0.89 &0.93 &0.86 &0.93 &0.89 \\
  &UTKF &0.94 &0.96 &0.91 &0.96 &0.93\\
  &UTKF-MC &0.85 &0.82 &0.91 &0.80 &0.86 \\
\hline
 RM & C-WF &0.87 &0.97 &0.76 &0.98 &0.85 \\
  &UTKF &0.86 &0.96 &0.75 &0.97 &0.84\\
  &UTKF-MC &0.73 &0.71 &0.75 &0.70 &0.73 \\
\hline
\end{tabular}
\caption{Performance of prototype-based representations for distinguishing syndromic facial images from controls. For each pooling method, logistic-regression classifiers were trained using 2,000 GMDB-Frequent images and 2,000 control images from either CASIA-WebFace or UTKFace. Performance was evaluated on a fixed syndromic cohort comprising the GMDB-Frequent test set and GMDB-Rare dataset (n=2,191), paired with held-out controls from CASIA-WebFace,  UTKFace, UTKFace match controls. \textbf{Legend:} Acc - Accuracy, PPV - Positive Predictive Value (aka precision), Sen - Sensitivity (aka recall), Spec - Specificity, GMP - Global Max Pooled, RM - Region Masked, C-WF - CASIA-WebFace, UTKF - UTKFace, UTKF-MC - UTKFace Mactched Controls }
\label{tab:binary_val}
\end{table}

\subsection{Syndrome identification performance}

We evaluated five CNN backbones(ResNet50, ResNet101 \cite{he2016},   IResNet100 \cite{deng2019}, ConvNeXt13, ConvNeXt26) to determine whether syndrome-identification performance and the effects of region masking were consistent across different feature extractors rather than specific to a single architecture. The number of prototypes ($P$) are each backbone's output channel count. Using the trained classification heads, IResNet100 achieved the strongest overall performance: region masking produced the highest Top-1 accuracy (33.56\%), whereas global max pooling produced the highest Top-5 and Top-10 accuracies (52.27\% and 59.86\%). Region masking improved Top-1 accuracy for four of the five backbones, although its effect on higher-rank accuracy varied.


\begin{table}[t]
\centering
\small
\begin{tabular}{lcccc}
\toprule
\multirow{2}{*}{\textbf{Model}} &
\multirow{2}{*}{\shortstack{\textbf{Number of}\\\textbf{prototypes}}} &
\multicolumn{3}{c}{\textbf{GMDB-Frequent}} \\
\cmidrule(lr){3-5}
& & \textbf{Top-1} & \textbf{Top-5} & \textbf{Top-10} \\
\midrule
RN50:GMP
    & \multirow{2}{*}{2048}
    & 21.8 & 34.1 & 40.9 \\
RN50:RM
    &
    & 23.0 & 31.2 & 37.9 \\
\midrule
RN101:GMP
    & \multirow{2}{*}{2048}
    & 7.0 & 17.1 & 25.2 \\
RN101:RM
    &
    & 4.8 & 14.5 & 20.3 \\
\midrule
IRN100:GMP
    & \multirow{2}{*}{512}
    & 33.3 & 52.3 & 59.9 \\
IRN100:RM
    &
    & 33.6 & 51.0 & 58.2 \\
\midrule
CNX13:GMP
    & \multirow{2}{*}{768}
    & 28.0 & 44.1 & 50.3 \\
CNX13:RM
    &
    & 32.0 & 45.7 & 52.6 \\
\midrule
CNX26:GMP
    & \multirow{2}{*}{768}
    & 30.6 & 45.8 & 52.0 \\
CNX26:RM
    &
    & 30.8 & 47.4 & 53.7 \\
\bottomrule
\end{tabular}
\caption{Syndrome-identification performance on the GMDB-Frequent test
set ($n=882$) using the trained PIP-Net classification head (all values are percents). The number
of prototypes is reported for each CNN backbone ($f_\theta$). Top-$1$,
Top-$5$, and Top-$10$ accuracies ($\mathrm{Acc}_k$) are reported for
global max-pooled and region-masked prototype representations across
five CNN backbones. \textbf{Legend:} RN50 - ResNet50, RN101 - ResNet101, IRN100 - IResNet100, CNX13 - ConvNeXt13, 
CNX26 - ConvNeXt26, GMP - Global Max Pooled, RM - Region Masked}
\label{tab:gmdb_syndrome_results_classification}
\end{table}

\begin{figure*}[t]
    \centering  \includegraphics[width=\textwidth]{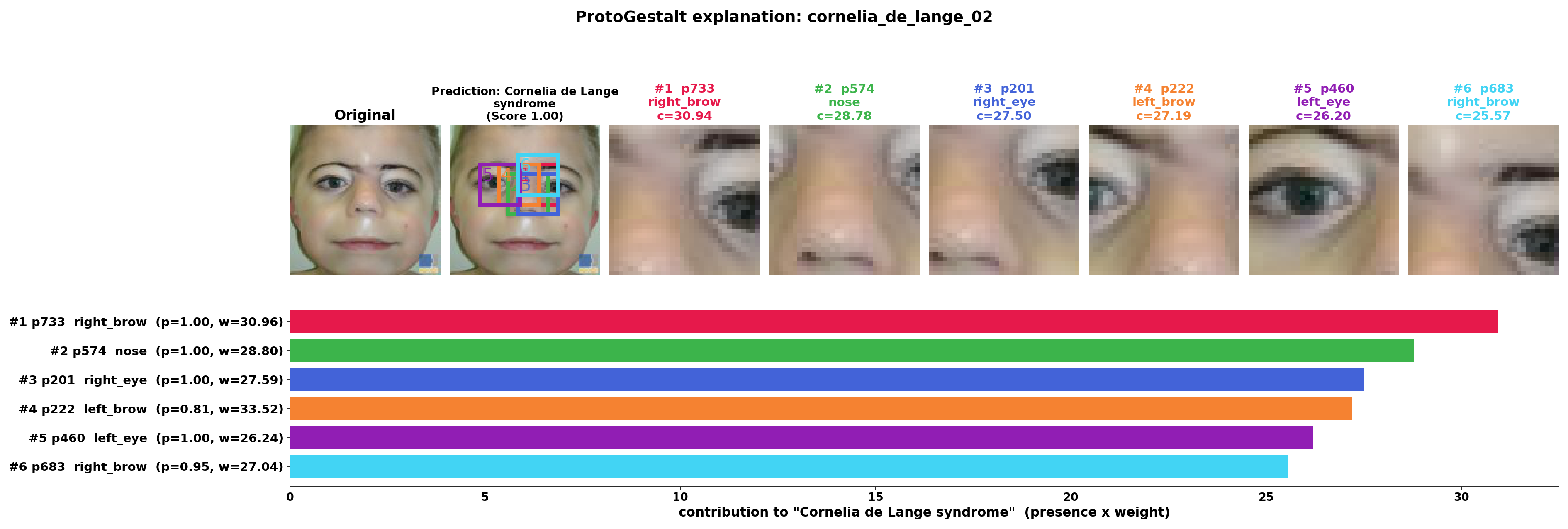}

    \caption{Local explanation of a prediction for a case of Cornelia de Lange syndrome. Prototypes are ranked by their contribution to the predicted class (contribution(c) = prototype presence value($\mathbf{p}$) × its classifier weight($\mathbf{w}$)). For each top prototype, the patches show the region of the image where it was most strongly activated, and the bar chart shows the corresponding contribution scores. The synthetic image shown in the illustration was generated with GestaltGAN for use in this publication.}
    \label{fig:prototypes}
\end{figure*}

\subsection*{Case study: Prototype consistency in Cornelia de Lange syndrome}

To assess the anatomical and clinical relevance of the learned prototypes, we selected
Cornelia de Lange syndrome (CdLS) as a case study and examined the consistency of the
prototypes driving its classification. The model (RM) correctly classified 30 of the 43 CdLS
test images, corresponding to a top-1 accuracy of 69.8\%. For each image, we identified
the ten prototypes that contributed most strongly to the CdLS prediction. These
decision-driving prototypes were consistent across patients, with a mean pairwise
Jaccard similarity of 0.335, compared with a random-subset baseline of 0.007. The most
frequently recurring prototypes localized to the eyebrows, eyes, nose, and cheeks
(Table~\ref{tab:cdls_prototypes}), which correspond to facial regions containing
features that are clinically relevant to the diagnosis of CdLS~\cite{Rohatgi2010}.

Qualitative analysis of the local explanations further showed that image patches
associated with the highest-contributing prototypes appear to capture characteristic known CdLS
facial phenotypes (Figure~\ref{fig:prototypes}). Specifically, prototype \#p683
appears to corresponded to synophrys (HP:0000664), \#p733 to thick eyebrows (HP:0000574),
\#p460 to long eyelashes (HP:0000343), and \#p574 to a short nose
(HP:0003196)~\cite{Kline2018}, though we note these are qualitative assessments. 
Together, these findings suggest that the model's
predictions are reproducible across patients and grounded in clinically meaningful
facial features rather than patient-specific or idiosyncratic image characteristics.

We extended this analysis to the 20 most frequent syndromes, for which the mean
top-1 accuracy was 50.1\%. Across these syndromes, the decision-driving prototypes
also remained consistent among patients, with a mean pairwise Jaccard similarity of
0.27, compared with a random-subset baseline of 0.007.

\begin{table}[t]
  \centering
  \begin{tabular}{llcc}
    \toprule
    Prototype & Region & Patients (of 43) & Fraction \\
    \midrule
    p683 & right\_brow  & 27 & 0.63 \\
    p733 & right\_brow  & 27 & 0.63 \\
    p574 & nose         & 27 & 0.63 \\
    p222 & left\_brow   & 26 & 0.60 \\
    p201 & right\_eye   & 25 & 0.58 \\
    p20  & left\_eye    & 24 & 0.56 \\
    p212 & left\_brow   & 24 & 0.56 \\
    p460 & left\_eye    & 23 & 0.53 \\
    p185 & left\_cheek  & 23 & 0.53 \\
    p755 & left\_cheek  & 21 & 0.49 \\
    \bottomrule
  \end{tabular}
  \caption{Prototypes most consistently driving the Cornelia de Lange syndrome
  prediction, ranked by the fraction of the 43 patients in whose top-10 contribution
  they appear.}
  \label{tab:cdls_prototypes}
\end{table}

\section{Discussion}
ProtoGestalt addresses a clinically broader task than models optimized solely for syndrome classification: it supports both syndromic-versus-control screening and syndrome identification while providing anatomically localized, additive prototype-based explanations. To situate this second capability against prior work, we additionally evaluated the prototype representations under the retrieval protocol of \citet{Hustinx2022}, covering both syndromes seen during training (GMDB-Frequent) and ultra-rare syndromes absent from training (GMDB-Rare); full results are reported in Technical Supplement: Appendix~A. ProtoGestalt did not match the syndrome-identification performance of specialized facial-phenotype embedding models, but its representations retained discriminative information for both seen and unseen syndromes while also separating syndromic images from controls. This trade-off may be valuable in front-line settings, where the initial clinical question is often whether a patient warrants genetic evaluation rather than which specific syndrome is present.

Region masking improved anatomical consistency and interpretability but reduced recall or overall predictive performance in several experiments. This finding suggests that strict spatial constraints may exclude informative global facial cues or interactions among facial regions. A further limitation is that individuals in CASIA-WebFace and UTKFace were assumed not to have rare disorders because their clinical status was unavailable. These datasets may therefore include undiagnosed individuals, potentially biasing estimates of screening performance. We also trained a simple logistic regression (LR) model using region-masked (RM) prototype representations from equal subsets of CASIA-WebFace and UTKFace to predict the source dataset of each image. The model achieved an AUPRC of 0.8, suggesting the presence of detectable dataset-specific signals among the control images. Consequently, the use of unmatched or naively assembled control datasets may inflate apparent screening performance by allowing models to exploit domain differences rather than clinically meaningful phenotypic features. Finally, prototype consistency and clinical relevance were evaluated only for Cornelia de Lange syndrome because of the scope of this study, limiting the generalizability of these findings to other disorders. Future work should therefore include a comprehensive cross-syndrome evaluation of prototype consistency and clinical relevance. It should also investigate graph neural networks applied to landmark-based facial graphs to model relationships among anatomical regions, with the goal of improving prototype quality, consistency, and diagnostic performance.

\section{Conclusion}

ProtoGestalt demonstrates that anatomically constrained prototypes can support interpretable facial phenotyping of rare genetic disorders. Its prototype-based representations also support  identification of disorders while providing localized facial evidence that can be inspected alongside model outputs. Unlike methods that rely primarily on post-hoc explanations, ProtoGestalt provides intrinsic, anatomically grounded evidence for its predictions. This framework may therefore serve as a foundation for more transparent and clinically useful facial-phenotyping systems.




\clearpage
\bibliography{aaai2027}

@article{Benichou2025,
   author = {Ludovic Benichou and Luan Breton and Nicolas Garcelon and Benjamin Benichou and Olivier Lienhard and Jeanne Amiel and Thomas Bongibault and Ana Julia Bravo Hidalgo and Valérie Cormier-Daire and Stanislas Lyonnet and Arnaud Picard and Marlène Rio and Ahmed Zaiter and Roman H. Khonsari and Quentin Hennocq},
   doi = {10.1002/ajmg.a.64298},
   issn = {15524833},
   issue = {3},
   journal = {American Journal of Medical Genetics, Part A},
   month = {3},
   pages = {685-694},
   pmid = {41194463},
   publisher = {John Wiley and Sons Inc},
   title = {Evaluation of a New Inclusive Next-Generation Synthetic Face Tool for Dysmorphology},
   volume = {200},
   url = {/doi/pdf/10.1002/ajmg.a.64298 https://onlinelibrary.wiley.com/doi/abs/10.1002/ajmg.a.64298 https://onlinelibrary.wiley.com/doi/10.1002/ajmg.a.64298},
   year = {2025}
}

@article{Khler2021,
   author = {Sebastian Köhler and Michael Gargano and Nicolas Matentzoglu and Leigh C. Carmody and David Lewis-Smith and Nicole A. Vasilevsky and Daniel Danis and Ganna Balagura and Gareth Baynam and Amy M. Brower and Tiffany J. Callahan and Christopher G. Chute and Johanna L. Est and Peter D. Galer and Shiva Ganesan and Matthias Griese and Matthias Haimel and Julia Pazmandi and Marc Hanauer and Nomi L. Harris and Michael J. Hartnett and Maximilian Hastreiter and Fabian Hauck and Yongqun He and Tim Jeske and Hugh Kearney and Gerhard Kindle and Christoph Klein and Katrin Knoflach and Roland Krause and David Lagorce and Julie A. McMurry and Jillian A. Miller and Monica C. Munoz-Torres and Rebecca L. Peters and Christina K. Rapp and Ana M. Rath and Shahmir A. Rind and Avi Z. Rosenberg and Michael M. Segal and Markus G. Seidel and Damian Smedley and Tomer Talmy and Yarlalu Thomas and Samuel A. Wiafe and Julie Xian and Zafer Yüksel and Ingo Helbig and Christopher J. Mungall and Melissa A. Haendel and Peter N. Robinson},
   doi = {10.1093/NAR/GKAA1043},
   issn = {1362-4962},
   issue = {D1},
   journal = {Nucleic acids research},
   month = {1},
   pages = {D1207-D1217},
   pmid = {33264411},
   publisher = {Nucleic Acids Res},
   title = {The Human Phenotype Ontology in 2021},
   volume = {49},
   url = {https://pubmed.ncbi.nlm.nih.gov/33264411/},
   year = {2021}
}

@article{Mao2025,
   author = {Xiaohao Mao and Yu Huang and Ye Jin and Lun Wang and Xuanzhong Chen and Honghong Liu and Xinglin Yang and Haopeng Xu and Xiaodong Luan and Ying Xiao and Siqin Feng and Jiahao Zhu and Xuegong Zhang and Rui Jiang and Shuyang Zhang and Ting Chen},
   doi = {10.1038/S41746-025-01452-1;SUBJMETA=139,692,699,700;KWRD=DIAGNOSIS,DISEASES},
   issn = {23986352},
   issue = {1},
   journal = {npj Digital Medicine},
   month = {12},
   pages = {1-15},
   publisher = {Nature Research},
   title = {A phenotype-based AI pipeline outperforms human experts in differentially diagnosing rare diseases using EHRs},
   volume = {8},
   url = {https://www.nature.com/articles/s41746-025-01452-1},
   year = {2025}
}

@article{Alsentzer2025,
   author = {Emily Alsentzer and Michelle M. Li and Shilpa N. Kobren and Ayush Noori and Isaac S. Kohane and Marinka Zitnik},
   doi = {10.1038/s41746-025-01749-1},
   issn = {23986352},
   issue = {1},
   journal = {npj Digital Medicine 2025 8:1},
   month = {6},
   pages = {380-},
   publisher = {Nature Publishing Group},
   title = {Few shot learning for phenotype-driven diagnosis of patients with rare genetic diseases},
   volume = {8},
   url = {https://www.nature.com/articles/s41746-025-01749-1},
   year = {2025}
}

@article{Li2019,
   author = {Qigang Li and Keyan Zhao and Carlos D. Bustamante and Xin Ma and Wing H. Wong},
   doi = {10.1038/s41436-019-0439-8},
   issn = {1530-0366},
   issue = {9},
   journal = {Genetics in Medicine 2019 21:9},
   month = {1},
   pages = {2126-2134},
   pmid = {30675030},
   publisher = {Nature Publishing Group},
   title = {Xrare: a machine learning method jointly modeling phenotypes and genetic evidence for rare disease diagnosis},
   volume = {21},
   url = {https://www.nature.com/articles/s41436-019-0439-8},
   year = {2019}
}

@article{Robinson2020,
   author = {Peter N. Robinson and Vida Ravanmehr and Julius O.B. Jacobsen and Daniel Danis and Xingmin Aaron Zhang and Leigh C. Carmody and Michael A. Gargano and Courtney L. Thaxton and Guy Karlebach and Justin Reese and Manuel Holtgrewe and Sebastian Köhler and Julie A. McMurry and Melissa A. Haendel and Damian Smedley},
   doi = {10.1016/j.ajhg.2020.06.021},
   issn = {15376605},
   issue = {3},
   journal = {American Journal of Human Genetics},
   month = {9},
   pages = {403-417},
   publisher = {Cell Press},
   title = {Interpretable Clinical Genomics with a Likelihood Ratio Paradigm},
   volume = {107},
   url = {http://www.cell.com/article/S0002929720302305/fulltext http://www.cell.com/article/S0002929720302305/abstract https://www.cell.com/ajhg/abstract/S0002-9297(20)30230-5},
   year = {2020}
}

@article{Ediae2025,
   author = {Grace U. Ediae and Alexandre White-Brown and Caitlin Chisholm and Ivan Terekhov and Jon Seymour and Jeff Guo and Nicholas Mitsakakis and Sarah L. Sawyer and Kym M. Boycott},
   doi = {10.1016/j.gim.2025.101570},
   issn = {15300366},
   issue = {11},
   journal = {Genetics in Medicine},
   month = {11},
   pages = {101570},
   pmid = {40856103},
   publisher = {Elsevier},
   title = {ThinkRare: A search algorithm to identify patients with undiagnosed rare genetic disease in an electronic medical record},
   volume = {27},
   url = {https://doi.org/10.1038/s41431-019-0508-0},
   year = {2025}
}

@article{Bhasuran2025,
   author = {Balu Bhasuran and Katharina Schmolly and Yuvraaj Kapoor and Nanditha Lakshmi Jayakumar and Raymond Doan and Jigar Amin and Stephen Meninger and Nathan Cheng and Robert Deering and Karl Anderson and Simon W. Beaven and Bruce Wang and Vivek A. Rudrapatna},
   doi = {10.1093/jamia/ocae141},
   issn = {1527974X},
   issue = {1},
   journal = {Journal of the American Medical Informatics Association},
   month = {1},
   pages = {63-70},
   pmid = {38946554},
   publisher = {Oxford Academic},
   title = {Reducing diagnostic delays in acute hepatic porphyria using health records data and machine learning},
   volume = {32},
   url = {https://dx.doi.org/10.1093/jamia/ocae141},
   year = {2025}
}

@article{Budych2012,
   author = {Karolina Budych and Thomas M. Helms and Carsten Schultz},
   doi = {10.1016/j.healthpol.2012.02.018},
   issn = {01688510},
   issue = {2-3},
   journal = {Health Policy},
   month = {5},
   pages = {154-164},
   pmid = {22464590},
   publisher = {Elsevier},
   title = {How do patients with rare diseases experience the medical encounter? Exploring role behavior and its impact on patient–physician interaction},
   volume = {105},
   url = {https://doi.org/10.1038/scientificamerican0872-66},
   year = {2012}
}

@article{Rohani-Montez2023,
   author = {S. Christy Rohani-Montez and Jennifer Bomberger and Cong Zhang and Jacob Cohen and Lucy McKay and William R.H. Evans},
   doi = {10.1016/j.gimo.2023.100808},
   issn = {29497744},
   issue = {1},
   journal = {Genetics in Medicine Open},
   month = {1},
   publisher = {Elsevier B.V.},
   title = {Educational needs in diagnosing rare diseases: A multinational, multispecialty clinician survey},
   volume = {1},
   url = {https://www.gimopen.org/action/showFullText?pii=S2949774423008178 https://www.gimopen.org/action/showAbstract?pii=S2949774423008178 https://www.gimopen.org/article/S2949-7744(23)00817-8/abstract},
   year = {2023}
}

@article{Molster2016,
author = {Caron Molster and Debra Urwin and Louisa Di Pietro and Megan Fookes and Dianne Petrie and Sharon Van Der Laan and Hugh Dawkins},
   doi = {10.1186/S13023-016-0409-Z/TABLES/9},
   issn = {17501172},
   issue = {1},
   journal = {Orphanet Journal of Rare Diseases},
   month = {3},
   pages = {1-12},
   pmid = {27012247},
   publisher = {BioMed Central Ltd.},
   title = {Survey of healthcare experiences of Australian adults living with rare diseases},
   volume = {11},
   url = {https://ojrd.biomedcentral.com/articles/10.1186/s13023-016-0409-z},
   year = {2016}
}

@article{Heuyer2017,
   author = {Thomas Heuyer and Sonia Pavan and Christine Vicard},
   doi = {10.2147/PROM.S131033},
   journal = {Patient Related Outcome Measures},
   month = {9},
   pages = {97-110},
   publisher = {Dove Medical Press Ltd.},
   title = {The health and life path of rare disease patients: results of the 2015 French barometer},
   volume = {Volume 8},
   url = {https://www.tandfonline.com/action/journalInformation?journalCode=dpro20},
   year = {2017}
}

@inproceedings{nauta2023,
  title     = {{PIP-Net}: Patch-Based Intuitive Prototypes for Interpretable Image Classification},
  author    = {Nauta, Meike and Schl{\"o}tterer, J{\"o}rg and van Keulen, Maurice and Seifert, Christin},
  booktitle = {Proceedings of the IEEE/CVF Conference on Computer Vision and Pattern Recognition (CVPR)},
  pages     = {2744--2753},
  year      = {2023}
}

@article{Wei2024,
   author = {Hang Wei and Lin Gao and Shuai Wu and Yina Jiang and Bin Liu},
   doi = {10.1093/BIOINFORMATICS/BTAE306},
   issn = {13674811},
   issue = {5},
   journal = {Bioinformatics},
   month = {5},
   pmid = {38715444},
   publisher = {Oxford Academic},
   title = {DiSMVC: a multi-view graph collaborative learning framework for measuring disease similarity},
   volume = {40},
   url = {https://dx.doi.org/10.1093/bioinformatics/btae306},
   year = {2024}
}

@article{Marwaha2022,
   author = {Shruti Marwaha and Joshua W Knowles and Euan A Ashley},
   doi = {10.1186/s13073-022-01026-w},
   issn = {1756-994X},
   issue = {1},
   journal = {Genome medicine},
   pages = {23},
   pmid = {35220969},
   publisher = {BioMed Central Ltd},
   title = {A guide for the diagnosis of rare and undiagnosed disease: beyond the exome},
   volume = {14},
   year = {2022}
}

@inproceedings{Akiba2019,
   author = {Takuya Akiba and Shotaro Sano and Toshihiko Yanase and Takeru Ohta and Masanori Koyama},
   booktitle = {Proceedings of the 25th ACM SIGKDD International Conference on Knowledge Discovery and Data Mining},
   title = {Optuna: A Next-generation Hyperparameter Optimization Framework},
   year = {2019}
}

@article{Chen2019,
   author = {Xiaoyi Chen and Nicolas Garcelon and Antoine Neuraz and Katy Billot and Marc Lelarge and Thomas Bonald and Hugo Garcia and Yoann Martin and Vincent Benoit and Marc Vincent and Hassan Faour and Maxime Douillet and Stanislas Lyonnet and Sophie Saunier and Anita Burgun},
   doi = {10.1016/j.jbi.2019.103308},
   issn = {15320464},
   journal = {Journal of Biomedical Informatics},
   month = {12},
   pmid = {31622800},
   publisher = {Academic Press Inc.},
   title = {Phenotypic similarity for rare disease: Ciliopathy diagnoses and subtyping},
   volume = {100},
   url = {https://pubmed.ncbi.nlm.nih.gov/31622800/},
   year = {2019}
}

@article{Gurovich2019,
   author = {Yaron Gurovich and Yair Hanani and Omri Bar and Guy Nadav and Nicole Fleischer and Dekel Gelbman and Lina Basel-Salmon and Peter M. Krawitz and Susanne B. Kamphausen and Martin Zenker and Lynne M. Bird and Karen W. Gripp},
   doi = {10.1038/s41591-018-0279-0},
   issn = {1546170X},
   issue = {1},
   journal = {Nature Medicine},
   month = {1},
   pages = {60-64},
   pmid = {30617323},
   publisher = {Nature Publishing Group},
   title = {Identifying facial phenotypes of genetic disorders using deep learning},
   volume = {25},
   year = {2019}
}

@article{Kirchhoff2025,
   author = {Aron Kirchhoff and Alexander Hustinx and Behnam Javanmardi and Tzung Chien Hsieh and Fabian Brand and Fabio Hellmann and Silvan Mertes and Elisabeth André and Shahida Moosa and Thomas Schultz and Benjamin D. Solomon and Peter Krawitz},
   doi = {10.1038/s41431-025-01787-z},
   issn = {14765438},
   issue = {3},
   journal = {European Journal of Human Genetics},
   month = {3},
   pages = {377-382},
   pmid = {39815041},
   publisher = {Springer Nature},
   title = {GestaltGAN: synthetic photorealistic portraits of individuals with rare genetic disorders},
   volume = {33},
   year = {2025}
}

@article{Ferry2014,
   author = {Quentin Ferry and Julia Steinberg and Caleb Webber and David R FitzPatrick and Chris P Ponting and Andrew Zisserman and Christoffer Nellåker},
   doi = {10.7554/elife.02020},
   issn = {2050-084X},
   journal = {eLife},
   month = {6},
   pmid = {24963138},
   publisher = {eLife Sciences Publications, Ltd},
   title = {Diagnostically relevant facial gestalt information from ordinary photos},
   volume = {3},
   year = {2014}
}

@article{Hsieh2022,
   author = {Tzung Chien Hsieh and Aviram Bar-Haim and Shahida Moosa and Nadja Ehmke and Karen W. Gripp and Jean Tori Pantel and Magdalena Danyel and Martin Atta Mensah and Denise Horn and Stanislav Rosnev and Nicole Fleischer and Guilherme Bonini and Alexander Hustinx and Alexander Schmid and Alexej Knaus and Behnam Javanmardi and Hannah Klinkhammer and Hellen Lesmann and Sugirthan Sivalingam and Tom Kamphans and Wolfgang Meiswinkel and Frédéric Ebstein and Elke Krüger and Sébastien Küry and Stéphane Bézieau and Axel Schmidt and Sophia Peters and Hartmut Engels and Elisabeth Mangold and Martina Kreiß and Kirsten Cremer and Claudia Perne and Regina C. Betz and Tim Bender and Kathrin Grundmann-Hauser and Tobias B. Haack and Matias Wagner and Theresa Brunet and Heidi Beate Bentzen and Luisa Averdunk and Kimberly Christine Coetzer and Gholson J. Lyon and Malte Spielmann and Christian P. Schaaf and Stefan Mundlos and Markus M. Nöthen and Peter M. Krawitz},
   doi = {10.1038/s41588-021-01010-x},
   issn = {15461718},
   issue = {3},
   journal = {Nature Genetics},
   month = {3},
   pages = {349-357},
   pmid = {35145301},
   publisher = {Nature Research},
   title = {GestaltMatcher facilitates rare disease matching using facial phenotype descriptors},
   volume = {54},
   year = {2022}
}

@article{Hustinx2022,
   author = {Alexander Hustinx and Fabio Hellmann and Ömer Sümer and Behnam Javanmardi and Elisabeth André and Peter Krawitz and Tzung-Chien Hsieh},
   doi = {10.1109/WACV56688.2023.00499},
   journal = {Proceedings - 2023 IEEE Winter Conference on Applications of Computer Vision, WACV 2023},
   month = {11},
   pages = {5007-5017},
   publisher = {Institute of Electrical and Electronics Engineers Inc.},
   title = {Improving Deep Facial Phenotyping for Ultra-rare Disorder Verification Using Model Ensembles},
   url = {http://arxiv.org/abs/2211.06764 http://dx.doi.org/10.1109/WACV56688.2023.00499},
   year = {2022}
}

@article{Halley2022,
   author = {Meghan C. Halley and Hadley Stevens Smith and Euan A. Ashley and Aaron J. Goldenberg and Holly K. Tabor},
   doi = {10.1038/S41588-022-01027-W},
   issn = {15461718},
   issue = {3},
   journal = {Nature genetics},
   month = {3},
   pages = {219},
   pmid = {35256804},
   publisher = {Nature Research},
   title = {A call for an integrated approach to improve efficiency, equity and sustainability in rare disease research in the United States},
   volume = {54},
   url = {https://pmc.ncbi.nlm.nih.gov/articles/PMC9016354/},
   year = {2022}
}

@article{Yi2014,
   author = {Dong Yi and Zhen Lei and Shengcai Liao and Stan Z. Li},
   month = {11},
   title = {Learning Face Representation from Scratch},
   url = {https://arxiv.org/pdf/1411.7923},
   year = {2014}
}

@article{Zhang2017,
   author = {Zhifei Zhang and Yang Song and Hairong Qi},
   doi = {10.1109/CVPR.2017.463},
   isbn = {9781538604571},
   journal = {Proceedings - 30th IEEE Conference on Computer Vision and Pattern Recognition, CVPR 2017},
   month = {2},
   pages = {4352-4360},
   publisher = {Institute of Electrical and Electronics Engineers Inc.},
   title = {Age Progression/Regression by Conditional Adversarial Autoencoder},
   volume = {2017-January},
   url = {https://arxiv.org/pdf/1702.08423},
   year = {2017}
}

@Article{pantel2020,
author="Pantel, Jean Tori
and Hajjir, Nurulhuda
and Danyel, Magdalena
and Elsner, Jonas
and Abad-Perez, Angela Teresa
and Hansen, Peter
and Mundlos, Stefan
and Spielmann, Malte
and Horn, Denise
and Ott, Claus-Eric
and Mensah, Martin Atta",
title="Efficiency of Computer-Aided Facial Phenotyping (DeepGestalt) in Individuals With and Without a Genetic Syndrome: Diagnostic Accuracy Study",
journal="J Med Internet Res",
year="2020",
month="Oct",
day="22",
volume="22",
number="10",
pages="e19263",
issn="1438-8871",
doi="10.2196/19263",
url="http://www.jmir.org/2020/10/e19263/",
url="https://doi.org/10.2196/19263",
url="http://www.ncbi.nlm.nih.gov/pubmed/33090109"
}

@inproceedings{grishchenko2020,
  title={Attention mesh: High-fidelity face mesh prediction in real-time},
  author={Grishchenko, Ivan and Ablavatski, Artsiom and Kartynnik, Yury and Raveendran, Karthik and Grundmann, Matthias},
  journal={arXiv preprint arXiv:2006.10962},
  year={2020}
}

@article{yazdani2025,
title = {Explainable artificial intelligence for pneumonia classification: Clinical insights into deformable prototypical part network in pediatric chest x-ray images},
journal = {Journal of Medical Imaging and Radiation Sciences},
volume = {56},
number = {5},
pages = {102023},
year = {2025},
issn = {1939-8654},
doi = {https://doi.org/10.1016/j.jmir.2025.102023},
url = {https://www.sciencedirect.com/science/article/pii/S1939865425001729},
author = {Elmira Yazdani and Aryan Neizehbaz and Najme Karamzade-Ziarati and Saeed Reza Kheradpisheh}
}

@misc{nauta2023_b,
      title={Interpreting and Correcting Medical Image Classification with PIP-Net}, 
      author={Meike Nauta and Johannes H. Hegeman and Jeroen Geerdink and Jörg Schlötterer and Maurice van Keulen and Christin Seifert},
      year={2023},
      eprint={2307.10404},
      archivePrefix={arXiv},
      primaryClass={cs.CV},
      url={https://arxiv.org/abs/2307.10404}, 
}

@article{gao2024,
author = {Gao, Yuchan and Su, Jing and Li, Jia and Wang, Shenglong and Li, Chao},
title = {A neural network framework based on ConvNeXt for side-channel hardware Trojan detection},
journal = {ETRI Journal},
volume = {47},
number = {2},
pages = {338-349},
doi = {https://doi.org/10.4218/etrij.2023-0448},
url = {https://onlinelibrary.wiley.com/doi/abs/10.4218/etrij.2023-0448},
eprint = {https://onlinelibrary.wiley.com/doi/pdf/10.4218/etrij.2023-0448},
year = {2025}
}

@inproceedings{he2016,
  title={Deep residual learning for image recognition},
  author={He, Kaiming and Zhang, Xiangyu and Ren, Shaoqing and Sun, Jian},
  booktitle={Proceedings of the IEEE conference on computer vision and pattern recognition},
  pages={770--778},
  year={2016}
}

@inproceedings{deng2019,
  title={Arcface: Additive angular margin loss for deep face recognition},
  author={Deng, Jiankang and Guo, Jia and Xue, Niannan and Zafeiriou, Stefanos},
  booktitle={Proceedings of the IEEE/CVF conference on computer vision and pattern recognition},
  pages={4690--4698},
  year={2019}
}

@article{Rohatgi2010,
title = {Facial Diagnosis of Mild and Variant CdLS: Insights From a Dysmorphologist Survey},
author = {Rohatgi, Sarika and Clark, Dinah and Kline, Antonie D. and Jackson, Laird G. and Pie, Juan and Siu, Victoria and Ramos, Feliciano J. and Krantz, Ian D. and Deardorff, Matthew A.},
doi = {10.1002/ajmg.a.33441},
issn = {1552-4825},
journal = {American journal of medical genetics. Part A},
language = {eng},
number = {7},
pages = {1641 - 1653},
publisher = {Wiley},
volume = {152A},
year = {2010},
url = {https://europepmc.org/articles/pmc4133091},
note = {is peer reviewed},
}

@article{Hsieh2024,
  title     = {GestaltMatcher Database---A global reference for facial phenotypic variability in rare human diseases},
  author    = {Hsieh, Tzung-Chien and Lesmann, Hellen and Hustinx, Alexander and Moosa, Shahida and Marchi, Elaine and Caro Martin, Maria del Pilar and Abdelrazek, Ibrahim and Pantel, Jean Tori and Klinkhammer, Hannah and ten Hagen, Merle and others},
  journal   = {Nature Medicine / Research Square (Preprint/Network Reference)},
  year      = {2024},
  doi       = {10.21203/rs.3.rs-4438861/v1}
}

@article{Kline2018,
  author = {Kline, Antonie D. and Classic, Margaret and Deardorff, Matthew A. and Krantz, Ian D. and others},
  title = {Diagnosis and management of Cornelia de Lange syndrome: first international consensus statement},
  journal = {Nature Reviews Genetics},
  volume = {19},
  number = {10},
  pages = {649--666},
  year = {2018},
  month = {10},
  doi = {10.1038/s41576-018-0031-0},
  url = {https://doi.org/10.1038/s41576-018-0031-0}
}


\end{document}